\documentclass{article}

 \usepackage[preprint]{neurips_2025}

\usepackage[utf8]{inputenc} 
\usepackage[T1]{fontenc}    
\usepackage{hyperref}       
\usepackage{url}            
\usepackage{booktabs}       
\usepackage{amsfonts}       
\usepackage{nicefrac}       
\usepackage{microtype}      
\usepackage{xcolor}         
\usepackage{amsmath,amsfonts,bm} 
\usepackage{amssymb} 
\usepackage{tcolorbox} 
\usepackage{subcaption} 
\usepackage{graphicx}
\usepackage{tikz}
\usepackage[export]{adjustbox}

\definecolor{neural_color}{RGB}{200, 30, 30}
\definecolor{behavior_color}{RGB}{30, 30, 200}
\definecolor{state_color}{RGB}{30, 100, 100}
\definecolor{shared_color}{RGB}{30, 200, 30}
\definecolor{backward_state_color}{RGB}{0, 0, 139} 

\newcommand{\cneural}[1]{\begingroup\color{neural_color}#1\endgroup}
\newcommand{\cbehavior}[1]{\begingroup\color{behavior_color}#1\endgroup}
\newcommand{\cstate}[1]{\textcolor{state_color}{#1}}

\newcommand{\bstate}[1]{\textcolor{backward_state_color}{#1}}

\newcommand{\vect}[1]{\bm{#1}}

\newcommand{\ve}{\vect{e}}
\newcommand{\veps}{\vect{\epsilon}}
\newcommand{\vv}{\vect{v}}
\newcommand{\vw}{\vect{w}}
\newcommand{\vx}{\vect{x}}
\newcommand{\vy}{\vect{y}}
\newcommand{\vz}{\vect{z}}

\newcommand{\ek}{\ve{}_{k}}
\newcommand{\vt}[1]{\vv{}_{#1}}
\newcommand{\w}[1]{\vw{}_{#1}}
\newcommand{\x}[1]{\cstate{\vx{}_{#1}}}
\newcommand{\xs}[1]{\cstate{\vx{}^{s}_{#1}}}
\newcommand{\y}[1]{\cneural{\vy{}_{#1}}}
\newcommand{\z}[1]{\cbehavior{\vz{}_{#1}}}

\newcommand{\vb}[1]{\vv{}^{b}_{#1}}
\newcommand{\wb}[1]{\vw{}^{b}_{#1}}
\newcommand{\xb}[1]{\cstate{\vx{}^{b}_{#1}}}

\newcommand{\wh}[1]{\hat{\vw{}}_{#1}}
\newcommand{\xh}[1]{\cstate{\hat{\vx{}}_{#1}}}
\newcommand{\yh}[1]{\cneural{\hat{\vy{}}_{#1}}}
\newcommand{\zh}[1]{\cbehavior{\hat{\vz{}}_{#1}}}
\newcommand{\xhb}[1]{\bstate{\hat{\vx{}}_{#1}^b}}

\newcommand{\ytilde}[1]{\cneural{\tilde{\vy{}}_{#1}}}
\newcommand{\ztilde}[1]{\cbehavior{\tilde{\vz{}}_{#1}}}
\newcommand{\ztildehat}[1]{\cbehavior{\hat{\tilde{\vz{}}}_{#1}}}

\newcommand{\Ytilde}[1]{\cneural{\tilde{Y}_{#1}}}
\newcommand{\Ztilde}[1]{\cbehavior{\tilde{Z}_{#1}}}

\newcommand{\xksup}[1]{\cstate{\hat{\vx{}}_{k}^{(#1)}}}

\newcommand{\xknextsup}[1]{\cstate{\hat{\vx{}}_{k+1}^{(#1)}}}

\newcommand{\wk}{\w{k}}
\newcommand{\wksup}[1]{\wk^{(#1)}}

\newcommand{\epsk}{\veps{}_k}
\newcommand{\vk}{\vt{k}}

\newcommand{\wbk}{\wb{k}}
\newcommand{\vbk}{\vb{k}}
\newcommand{\epsbk}{\veps{}_k^b}

\newcommand{\yk}{\y{k}}
\newcommand{\zk}{\z{k}}

\newcommand{\Pt}{\tilde{P}}
\newcommand{\Nt}{\tilde{N}}

\newcommand{\Pxerr}[1]{P_{#1}}
\newcommand{\Px}[1]{\Pt_{#1}}

\newcommand{\PxRev}[1]{\Nt_{#1}}

\newcommand{\CovXs}{\Sigma_{x}}
\newcommand{\CovY}{\Sigma_{y}}
\newcommand{\CovXsY}{G_{y}}
\newcommand{\CovE}{\Sigma_{e}}
\newcommand{\CovXsZ}{G_{z}}

\newcommand{\Rsup}[1]{\mathbb{R}^{#1}}
\newcommand{\inR}[1]{\in \mathbb{R}^{#1}}

\title{Preferential subspace identification (PSID) with forward-backward smoothing}

\author{
Omid G.~Sani \\
\texttt{omidsani@gmail.com} \\
\And
Maryam M. Shanechi\thanks{Affiliations: Departments of Electrical and Computer Engineering (both authors), Biomedical Engineering (M.M.S.), and Computer Science (M.M.S.), University of Southern California, Los Angeles, CA, USA.} \\
\texttt{shanechi@usc.edu}
}

\begin{document}

\maketitle

\begin{abstract}
System identification methods for multivariate time-series, such as neural and behavioral recordings, have been used to build models for predicting one from the other. For example, Preferential Subspace Identification (PSID) builds a state-space model of a primary time-series (e.g., neural activity) to optimally predict a secondary time-series (e.g., behavior). However, PSID focuses on optimal prediction using \textit{past} primary data, even though in offline applications, better estimation can be achieved by incorporating concurrent data (filtering) or all available data (smoothing). Here, we extend PSID to enable optimal filtering and smoothing. \textit{First}, we show that the presence of a secondary signal makes it possible to uniquely identify a model with an optimal Kalman update step (to enable filtering) from a family of otherwise equivalent state-space models. Our filtering solution augments PSID with a reduced-rank regression step that directly learns the optimal gain required for the update step from data. We refer to this extension of PSID as \textit{PSID with filtering}. \textit{Second}, inspired by two-filter Kalman smoother formulations, we develop a novel forward-backward \textit{PSID smoothing} algorithm where we first apply PSID with filtering and then apply it again in the reverse time direction on the residuals of the filtered secondary signal. We validate our methods on simulated data, showing that our approach recovers the ground-truth model parameters for filtering, and achieves optimal filtering and smoothing decoding performance of the secondary signal that matches the ideal performance of the true underlying model. This work provides a principled framework for optimal linear filtering and smoothing in the two-signal setting, significantly expanding the toolkit for analyzing dynamic interactions in multivariate time-series.
\end{abstract}

\section{Introduction}

Given a time series $\y{k}$, system identification is the problem of finding a latent state space model that describes the second-order statistics of $\y{1}$ to $\y{N}$.

Given a state space model, the problem of \textit{prediction} is finding the optimal estimation of the state $\x{k}$ at a given time sample $k$ given all past samples of $\y{}$. \textit{Filtering} is the problem of estimating $\x{k}$ at a given time step $k$ given all samples of $\y{}$ up to and including the current time step $\y{k}$. Finally, \textit{smoothing} is the problem of estimating $\x{k}$ at a given time step given samples of $\y{}$ up to a future time step after $k$.

For models with directly measurable states, e.g., kinematics of an object, all three problems have unique solutions. For models with latent states, the exact state is ultimately an internal characteristic of the system and its alternative estimates are only preferable insofar as they can be validated against an observable external characteristic of the system \citep{katayamaSubspaceMethodsSystem2006}. For example, an estimation of the observation itself $\y{k}$ based on the estimated latent state gives one measurable way to evaluate the estimated latent state. However, while estimation of $\y{k}$ using its past values, i.e., \textit{prediction}, is non-trivial,  the \textit{filtering} and \textit{smoothing} of $\y{k}$ is not. Specifically, assuming zero-mean additive observation noises, the best \textit{estimation} of any given sample $\y{k}$ (in the sense of having minimum expected value of squared error) would simply be its observed value if that sample $\y{k}$ itself is observed. To confirm this statement, note that the expected squared error of such estimation (i.e., estimating some $x$ as the noisy measured $y=x+\epsilon$) would be the covariance of the additive noise, which is the fundamental minimum error possible. In this one-signal setting thus the filtering and smoothing problems have trivial solutions where the estimated value of $\y{k}$ is the measured $\y{k}$ itself.

Beyond the aforementioned scenario, a two-signal setup for system identification may be at hand, such as the one we discuss in \citet{saniModelingBehaviorallyRelevant2021}. In this scenario, both a primary time series $\yk{}$ and a secondary time series $\zk{}$ are available, and the objective is to learn the dynamics of $\yk{}$ while dissociating its dynamics that are related to $\zk{}$ from those that are not and prioritizing the former. Preferential Subspace Identification (PSID) \citep{saniModelingBehaviorallyRelevant2021} optimally finds these model parameters, but it has only been shown and validated in the setting of \textit{predicting} the secondary signal from past samples of the primary signal. Critically though, as we show here, the two-signal system identification scenario also enables \textit{filtering} and \textit{smoothing} of the secondary signal.

The contributions of this work are two-fold. First, we extend PSID to enable optimal \textit{filtering} of the secondary signal. Our solution involves deriving the optimal Kalman update step for the secondary signal using reduced rank regression on top of PSID. 
Second, we further extend PSID to the smoothing problem. For smoothing, we develop a solution inspired by the forward-backward filtering formulation for Kalman smoothing, where we apply our extended PSID with filtering in a forward pass and also a backward pass on the residual secondary signal. We validate our results in simulations.

\section{Methods}

This section details the development of our proposed methods. We first lay the groundwork by reviewing prerequisite concepts in state-space modeling: the Kalman filter and smoother, different model formulations, and core principles of model identifiability (Sections~\ref{sec:kalman_filter}-\ref{eq:system_identification}). Building on this foundation, we present our primary contributions (Section~\ref{sec:psid_main}), where we introduce our novel extensions for optimal filtering and smoothing with PSID. We conclude by outlining the simulation framework and metrics used for validation (Section~\ref{sec:evaluation_metrics}).

\subsection{Model formulation}
\label{sec:model_formulation}

We model the temporal dynamics of two time-series $\yk{} \in \Rsup{n_y}$ and $\zk{} \inR{n_z}$ in terms of the latent state $\xs{k} \inR{n_x}$ as 

\begin{equation}
   \label{eq:stochastic_form}
   \left\{
   \begin{array}{cccccc}
       \xs{k+1} & = & A   & \xs{k} &+& \wk{}
       \\
       \yk{}    & = &  C_y & \xs{k} &+& \vk{}
       \\
       \zk{}    & = & C_z & \xs{k} &+& \epsk{}
   \end{array}
   \right.
\end{equation}

where $\wk{} \inR{n_x}$ and $\vk{} \inR{n_y}$ are white Gaussian noises with the following cross-correlation: 

\begin{equation}
   \label{eq:def_qrs}
   \mathbb{E}\{ 
    \begin{bmatrix} \wk{} \\ \vk{} \end{bmatrix}
    \begin{bmatrix} \wk{} \\ \vk{} \end{bmatrix}^T
    \} = 
    \begin{bmatrix} 
        Q   & S \\ 
        S^T & R 
    \end{bmatrix}.
\end{equation}

\subsection{Kalman filter}
\label{sec:kalman_filter}
Given observations $\y{0}, \y{1}, \dots, \y{k}$, a Kalman filter gives the optimal (in the sense of having the minimum mean squared error) estimate of the latent state $\xs{k+1}$ as follows \citep{andersonOptimalFiltering2012,astromComputerControlledSystemsTheory2013}:
\begin{subequations}
\label{eq:Kalman_state}
\begin{equation}
   \label{eq:Kalman_state_update}
   \xh{k|k} = 
   \xh{k|k-1} + K_f (\y{k} - C_y \xh{k|k-1}) = 
   (A - K_f C_y) \xh{k|k-1} + K_f \y{k}
\end{equation}
\begin{equation}
   \label{eq:Kalman_state_prediction}
   \xh{k+1|k} = 
   A \xh{k|k} + \wh{k|k} = A \xh{k|k} + K_v (\y{k} - C_y \xh{k|k-1}) = 
   (A - K C_y) \xh{k|k-1} + K \y{k}
\end{equation}
\end{subequations}
where Kalman gains $K_f$, $K_v$ and $K$ are defined as
\begin{subequations}
\begin{equation}
   \label{eq:Kalman_Kf_def}
   K_f \triangleq \Pxerr{k|k-1} C_y^T (C_y \Pxerr{k|k-1} C_y^T + R)^{-1}
\end{equation}
\begin{equation}
   \label{eq:Kalman_Kv_def}
   K_v \triangleq S (C_y \Pxerr{k|k-1} C_y^T + R)^{-1}
\end{equation}
\begin{equation}
   \label{eq:Kalman_K_def}
   K \triangleq A K_f + K_v = (A \Pxerr{k|k-1} C_y^T + S) (C_y \Pxerr{k|k-1} C_y^T + R)^{-1}
\end{equation}
\label{eq:Kalman_gain_defs}
\end{subequations}
and $\Pxerr{k|k-1}$ represents the error covariance of the estimated state, defined as:
\begin{equation}
   \label{eq:Kalman_P_def}
   \Pxerr{k|k-1} \triangleq \mathbb{E}\left[ (\xh{k|k-1} - \xs{k})(\xh{k|k-1} - \xs{k})^T \right]
\end{equation}
This covariance follows the following recursive Riccati equations:
\begin{subequations}
\label{eq:Kalman_P}
\begin{equation}
   \label{eq:Kalman_P_update}
   \Pxerr{k|k} = 
   \Pxerr{k|k-1} - \Pxerr{k|k-1} C_y^T (C_y \Pxerr{k|k-1} C_y^T + R)^{-1} C_y \Pxerr{k|k-1} = \Pxerr{k|k-1} - K_f C_y \Pxerr{k|k-1}
\end{equation}
\begin{equation}
   \label{eq:Kalman_P_prediction}
   \begin{array}{cl}
   \Pxerr{k+1|k} & = 
   A \Pxerr{k|k-1} A^T + Q - (A \Pxerr{k|k-1} C_y^T + S) (C_y \Pxerr{k|k-1} C_y^T + R) (A \Pxerr{k|k-1} C_y^T + S)^T \\ [1em]
   & = A \Pxerr{k|k-1} A^T + Q - K (C_y \Pxerr{k|k-1} C_y^T + R)^{-1} K^T .
   \end{array}
\end{equation}
\end{subequations}

Initial conditions for the Kalman filter also need to be specified for the above recursive equations to start, but given their limited effect on the steady state performance of stable models, they can usually be chosen as
\begin{equation}
    \xh{0|-1} = \vect{0}, \qquad \Pxerr{0|-1} = I
\end{equation}
where $\xh{0|-1}$ is the initial state estimate and $\Pxerr{0|-1}$ is the initial error covariance.

For the stationary state space model of equation \ref{eq:stochastic_form}, when the Riccati equations have a stable solution, at steady state, $\Pxerr{k+1|k}$ and $\Pxerr{k|k}$ converge to steady state values that we denote by $\Pxerr{p}$ and $\Pxerr{}$, respectively. The steady state version of equations \ref{eq:Kalman_P} are thus
\begin{subequations}
\label{eq:Kalman_P_steady_state}
\begin{equation}
   \label{eq:Kalman_P_steady_state_update}
   \Pxerr{} = 
   \Pxerr{p} - \Pxerr{p} C_y^T (C_y \Pxerr{p} C_y^T + R)^{-1} C_y \Pxerr{p} 
\end{equation}
\begin{equation}
   \label{eq:Kalman_P_steady_state_prediction}
   \begin{array}{cl}
   \Pxerr{p} & = 
   A \Pxerr{p} A^T + Q - (A \Pxerr{p} C_y^T + S) (C_y \Pxerr{p} C_y^T + R) (A \Pxerr{p} C_y^T + S)^T 
   \end{array}.
\end{equation}
\end{subequations}

\subsection{Stochastic versus predictor form formulations}
\label{sec:stochastic_vs_predictor}

Having reviewed the Kalman filter, we can now discuss an important concept. Equations \ref{eq:stochastic_form}-\ref{eq:def_qrs} are only one of several equivalent ways to formulate the multivariate Gaussian random process $\y{k}$ as a latent state space model. Specifically, this formulation, which is repeated below, is referred to as the forward stochastic model \citep{vanoverscheeSubspaceIdentificationLinear1996}:

\begin{tcolorbox}[title=The stochastic form, halign title=center]
\centering
\begin{subequations}
\label{eq:stochastic_box}
\begin{align}
\xs{k+1} &= A \xs{k} + \wk{} \\
\y{k} &= C_y \xs{k} + \vk{} \\
\mathbb{E}\left(\begin{bmatrix} \w{p} \\ \vt{p} \end{bmatrix} \begin{bmatrix} \w{q} \\ \vt{q} \end{bmatrix}^T\right) &= \begin{pmatrix} Q & S \\ S^T & R \end{pmatrix} \delta_{pq}
\end{align}
\end{subequations}
\begin{subequations}
\label{eq:stochastic_box_covariances}
\begin{align}
\mathbb{E}[\xs{k} (\xs{k})^T] &\triangleq \CovXs = A \CovXs A^T + Q, \\
\mathbb{E}[\y{k} \y{k}^T] &\triangleq \CovY = C_y \CovXs C_y^T + R, \\
\mathbb{E}[\xs{k+1} \y{k}^T] &\triangleq \CovXsY = A \CovXs C_y^T + S.
\end{align}
\end{subequations}
\end{tcolorbox}

Here, equations \ref{eq:stochastic_box_covariances}a-c are obtained by taking covariances and cross covariances from equations \ref{eq:stochastic_box}a-b. These equations specify the relationship between the $Q$, $R$, and $S$ noise covariances with the latent state and observation covariances $\CovXs$, $\CovY$, and $\CovXsY$. Specifically, to find $\CovXs$, $\CovY$, and $\CovXsY$ based on the former, we can simply use equations \ref{eq:stochastic_box_covariances}a-c. Conversely, to find $Q$, $R$, and $S$ based on $\CovXs$, $\CovY$, and $\CovXsY$, we can solve the Lyapunov equation (equation \ref{eq:stochastic_box_covariances}a) to find a solution for $\CovXs$ and then replace that solution in equations \ref{eq:stochastic_box_covariances}b-c to find $R$ and $S$, respectively.

An alternative equivalent formulation that describes the exact same second order statistics for $\y{k}$ is the ``forward predictor form" formulation provided below, where the latent state $\x{k}$ is taken to be the Kalman estimated state, i.e., $\x{k} \triangleq \xh{k|k-1}$:

\begin{tcolorbox}[title=The predictor form, halign title=center]
\centering
\begin{subequations}
\label{eq:predictor_box}
\begin{align}
\x{k+1} &= A \x{k} + K \ek{} \\
\yk &= C_y \x{k} + \ek{}
\end{align}
\end{subequations}
\begin{subequations}
\label{eq:predictor_box_covariances}
\begin{align}
\mathbb{E}[\x{k} (\x{k})^T] &\triangleq \Px{k} \\
\mathbb{E}[\ek{} \ek{}] &\triangleq \CovE = \CovY - C_y \Px{k} C_y^T
\end{align}
\begin{align}
\Px{k} = A \Px{k-1} A^T &+ (\CovXsY - A \Px{k-1} C_y^T)(\CovY - C_y \Px{k-1} C_y^T)^{-1}(\CovXsY - A \Px{k-1} C_y^T)^T \\
K_{k-1} &= (\CovXsY - A \Px{k-1} C_y^T)(\CovY - C_y \Px{k-1} C_y^T)^{-1}
\end{align}
\end{subequations}
\end{tcolorbox}

Here, $\ek{}$ is the part of the observation $\y{k}$ that is not predictable from past observation samples. $\ek{}$ is also known as the innovation, which is why this formulation is known as the innovation form. Notably, simply replacing $\ek{}$ in equation \ref{eq:predictor_box}a with its definition from equation \ref{eq:predictor_box}b (i.e., $\y{k} - C_y \x{k}$) yields the Kalman filter equation \ref{eq:Kalman_state_prediction}. After that replacement, this formulation is known as the predictor form. Hereafter, for simplicity, we refer to both of these closely related formulations (innovation and predictor forms) as the predictor form.

Equation \ref{eq:predictor_box_covariances}a defines the covariance of the Kalman predicted state (i.e., $\x{k}$) itself, which is different from the error covariance of the predicted state (i.e., $\Pxerr{k}$). The relation of these two covariances can be derived by taking covariance from the relation between the underlying entities \citep{vanoverscheeSubspaceIdentificationLinear1996}:
\begin{subequations}
\begin{equation}
   \xs{k} = \x{k} + (\xs{k} - \x{k}) \\ 
\end{equation}
\begin{equation}
   \CovXs = \Px{k} + \Pxerr{k}
\end{equation}
\end{subequations}
where we have used the fact that the Kalman prediction error ($\xs{k} - \x{k}$) is orthogonal to Kalman predicted state ($\x{k}$). Equation \ref{eq:predictor_box_covariances}b is obtained by taking covariance from equation \ref{eq:predictor_box}b. Equation \ref{eq:predictor_box_covariances}c is an equivalent formulation of the Riccati equation \ref{eq:Kalman_P_prediction}, related via equation \ref{eq:stochastic_box_covariances}. Finally, equation \ref{eq:predictor_box_covariances}d is an alternative equivalent formulation for the Kalman gain equation \ref{eq:Kalman_K_def}.

While both the stochastic and predictor forms generate the same second-order statistics for the observations $\y{k}$, they use different model parameters. It is straightforward to find the predictor form parameters given the stochastic form parameters by simply computing the Kalman filter parameters for the stochastic model (see equations \ref{eq:Kalman_gain_defs}-\ref{eq:Kalman_P}). This conversion is indeed unique (within a similarity transform, see section \ref{sec:similarity_transforms}) because each model has a specific unique Kalman filter associated with it.

The opposite conversion, from predictor form to stochastic form, is not unique and has an infinite number of solutions, even beyond similarity transforms. This is because the stochastic form is a redundant representation with more parameters than needed to describe the second-order statistics of the observations $\y{k}$ \citep{vanoverscheeSubspaceIdentificationLinear1996,katayamaSubspaceMethodsSystem2006}. The family of solutions for this conversion is given by Faurre's theorem \citep{vanoverscheeSubspaceIdentificationLinear1996}.

\textbf{Faurre's Theorem:} 
\label{sec: faurres_theorem}
The set of all state covariance matrices $\CovXs$ that generate the same output covariance statistics for $\y{k}$ is a closed, convex, and bounded set characterized by the inequality:
\begin{equation}
   \Px{} \leq \CovXs \leq \PxRev{}^{-1}
\end{equation}
where:
\begin{itemize}
    \item $\Px{}$ is the unique solution to the forward Riccati equation (equation \ref{eq:predictor_box_covariances}c),
    \item $\PxRev{}$ is the unique solution to the backward Riccati equation (see \citet{vanoverscheeSubspaceIdentificationLinear1996}),
    \item $\CovXs$ is the state covariance matrix for the stochastic form.
\end{itemize}

For every $\CovXs$ satisfying this inequality, the noise covariances for the stochastic form can be constructed by replacing $\CovXs$ in equation \ref{eq:stochastic_box_covariances}.

Thus, there are infinitely many stochastic models (with different $Q$, $R$, $S$) that generate the same second-order statistics for $\y{k}$, all parameterized by the choice of $\CovXs$ within the bounds above. 

The redundancy of the stochastic form in terms of model parameters can also be confirmed by simply counting the number of parameters for stochastic and predictor forms. The stochastic and predictor forms can be summarized with the set of parameters $\{A, C_y, Q, R, S\}$ and $\{A, C_y, K, \Sigma_e\}$, respectively. The $A$ and $C_y$ are shared between them, but the noises are described with $(n_x+n_y)(n_x+n_y+1)/2 = n_x^2/2 + n_x/2 + n_x n_y + n_y^2/2 + n_y/2$ (for $Q, R, S$) versus $n_x n_y + n_y^2/2 + n_y/2$ (for $K, \Sigma_e$) independent parameters (i.e., not counting complex conjugate terms), for stochastic versus predictor forms, respectively. As such, the stochastic form uses $n_x(n_x+1)/2$ more parameters to describe the same $\y{k}$.

Critically, as far as the time series $\y{k}$ on its own is concerned, all stochastic representations of the model are equivalent. A key insight presented in this work however is that this is no longer the case in the PSID setting, where a second time series $\z{k}$ is also measured during modeling. Before discussing that however, we will need to also review Kalman smoothing.

\subsection{Kalman smoother}
\label{sec:kalman_smoother}

Kalman smoothing provides the optimal estimate of the latent state $\xs{k}$ at time $k$ given \textit{all} observations up to the final time $N$, i.e., $\xh{k|N}$. This is in contrast to the Kalman filter, which provides the optimal estimate of $\xs{k}$ at time $k$ given observations up to $k$ ($\xh{k|k}$), and Kalman prediction, which estimates the next state $\xs{k+1}$ given observations up to $k$ ($\xh{k+1|k}$). One widely used formulation for smoothing is the Rauch-Tung-Striebel (RTS) smoother, which we describe below.

\paragraph{RTS Smoother (Rauch-Tung-Striebel)}
In the RTS smoother \citep{rauchMaximumLikelihoodEstimates1965}, after the Kalman filter runs in the forward direction, a second estimation step runs in the reverse time direction on the data to update the Kalman filter state estimations based on all the observed future data. The backward estimation is also recursive and can be formulated as follows:
\begin{subequations}
\label{eq:RTS_smoother}
\begin{align}
    L_k &= \Pxerr{k|k} A^T (\Pxerr{k+1|k})^{-1} \\
    \Pxerr{k|N} &= \Pxerr{k|k} + L_k (\Pxerr{k+1|N} - \Pxerr{k+1|k}) L_k^T \\
    \xh{k|N} &= \xh{k|k} + L_k (\xh{k+1|N} - A \xh{k|k})
\end{align}
\end{subequations}
where $\xh{k|N}$ and $\Pxerr{k|N}$ are the smoothed state estimate and covariance, respectively. Note that the ``initial'' state of this reverse estimation, i.e., $\xh{N|N}$, is the last filtered state from the forward pass so it is known when the backwards estimation starts.

A useful interpretation of the RTS smoother is that the smoothed state is a weighted average of the forward (filtered) and backward (smoothed) estimates:
\begin{equation}
    \xh{k|N} = (I - L_k A) \xh{k|k} + L_k \xh{k+1|N}
\end{equation}

Importantly, the backward recursive steps in the RTS formulation (equation \ref{eq:RTS_smoother}c) look like a filter, except they are not applied on observed data, $\y{k}$; rather, they are applied on the Kalman filter states, $\xh{k|k}$, which are the pseudo-observations of this backward filter. Is it possible to reformulate the Kalman smoothing problem as a forward and backward filtering problem where both filters are applied on the observed data, $\y{k}$? The answer is yes \citep{fraserOptimumLinearSmoother1969}.

\paragraph{Forward-backward (two-filter) smoother}

The same smoothed state estimates as the RTS smoother can also be obtained by combining a forward filter with a backward filter, in an approach called the two-filter or forward-backward smoother \citep{fraserOptimumLinearSmoother1969, kitagawaNoteRelationBalenzelas2023}. Importantly, the backward filter here is not the same as the backward recursion in the RTS smoother: it uses different state and covariance variables, which we denote with a superscript $b$.

\paragraph{Backward filter} As in the RTS formulation, the forward filter is a Kalman filter. The backward filter, proceeds from $N$ to $1$ and is defined as follows:

\subparagraph{Initial condition:}
\begin{equation}
\label{eq:two-filter_backwards_init}
   \xhb{N|N+1} = \vect{0}, \qquad P_{N|N+1}^b = 0
\end{equation}

\subparagraph{Update step:}
\begin{subequations}
\label{eq:backward_update}
\begin{align}
    \xhb{k|k} &= \xhb{k|k+1} + C_y^T R^{-1} \y{k} \\
    P_{k|k}^b &= P_{k|k+1}^b + C_y^T R^{-1} C_y
\end{align}
\end{subequations}

\subparagraph{Prediction step:}
\begin{subequations}
\label{eq:backward_prediction}
\begin{align}
    J_k &= P_{k|k}^b (P_{k|k}^b + Q^{-1})^{-1} \\
    \xhb{k-1|k} &= A^T (I - J_k) \xhb{k|k} \\
    P_{k-1|k}^b &= A^T (I - J_k) P_{k|k}^b A
\end{align}
\end{subequations}
Note that this formulation from \citep{kitagawaNoteRelationBalenzelas2023} assumes that there is no cross-correlation between the state and observation noises (i.e., $S = 0$). 


\subparagraph{Forward-backward weighted average smoother}
After running both the forward (Kalman) and backward filters, the smoothed state and covariance at each time $k$ can be computed as a weighted average:
\begin{align}
    \Pxerr{k|N} &= \left( \Pxerr{k|k}^{-1} + P_{k|k+1}^b \right)^{-1} \\
    \xh{k|N} &= \Pxerr{k|N} \Pxerr{k|k}^{-1} \xh{k|k} + \Pxerr{k|N} \xhb{k|k+1}
   \label{eq:forward_backward_smoother}
\end{align}
where $\xhb{k|k+1}$ and $P_{k|k+1}^b$ are the backward filter state and covariance, and $\xh{k|k}$ and $\Pxerr{k|k}$ are the forward (Kalman) filter state and covariance. Note that the backward filter is distinct from the RTS smoother variables, but yield the same optimal smoothed estimate $\xh{k|N}$.

Also note that although we use the notation $P_{k|k+1}^b$, this quantity is not the covariance of any quantity, rather it is the \textit{inverse} covariance of the backward filter, which is why it is initialized with $\vect{0}$ in equation \ref{eq:two-filter_backwards_init}. This alternative formulation for a Kalman filter that is based on inverse covariances is known as the \textit{information filter}. Nevertheless, unlike in the RTS formulation, in the two-filter formulation, the backward pass is applied to the observations, just like the forward pass. Finally, it is also worth noting that the backward filter in the forward-backward smoother formulation is different from the filter associated with the backward stochastic model \citep{vanoverscheeSubspaceIdentificationLinear1996} that is equivalent to equation \ref{eq:stochastic_form} (i.e., the backward system is a different model).

The forward-backward smoother formulation is notable because it is closely related to the method we develop in this work. Briefly, in the forward-backward smoother literature, the backward filter parameters are based on the state-space model parameters (as shown in equations \ref{eq:backward_update}-\ref{eq:backward_prediction}). In contrast, in this work, we learn both the forward and backward filter parameters from the data.

\subsection{Similarity transforms and equivalent models beyond them}
\label{sec:similarity_transforms}
Latent state space models are a fundamentally redundant representation, in the sense that one could write infinitely many different state space equations like equation \ref{eq:stochastic_form} that have different parameters but are equivalent and describe the exact same second order statistics for observation time series $\y{k}$ and $\z{k}$ \citep{vanoverscheeSubspaceIdentificationLinear1996,katayamaSubspaceMethodsSystem2006}.

Since the latent state $\x{k}$ is by definition not measured and does not correspond to any physical quantity all latent state space models that describe the statistics of the observed data (e.g., $\y{k}$) are equally valid, regardless of their exact latent state. In the one-signal setting, only $\y{k}$ is observed and thus all models that describe the same second-order statistics of $\y{k}$ (per Faurre's theorem) are equally valid. In the two-signal setting of PSID, only models are equally valid that further produce the same cross-correlative statistics for the two signals, which would mean that they yield similar conditional probability for $\z{k}$ given $\y{1:N}$. As we will show in this work, this allows us to narrow down the parameter space and find models that are optimal in prediction of $\z{k}$ using $\y{k}$.

\subsection{System identification and internal versus external characteristics of the model}
\label{eq:system_identification}
System identification or model fitting is the problem of finding a set of model parameter that represent a given training data well. As explained in section \ref{sec:stochastic_vs_predictor}, certain model representations are more redundant than others, meaning that there are more ways to describe the same data statistics using them. Specifically, the stochastic form latent state space model (equation \ref{eq:stochastic_box}) is a redundant representation, with infinitely many sets of $\{Q,R,S\}$ parameters giving the same second order statistics of $\y{k}$ (see Faurre's theorem). For this reason, the $\{Q,R,S\}$ parameters are not uniquely identifiable regardless of the method used for learning the model and the available training data. In other words, these parameters are internal characteristics of the stochastic form model and thus do not have a one-to-one manifestation on any measurable property of the system \citep{katayamaSubspaceMethodsSystem2006}. In contrast, the Kalman filter that is optimal for any stable Gaussian random process $\y{k}$ can uniquely be estimated, which is why the predictor form parameters are all external characteristics of the system and are thus uniquely identifiable (within a similarity transform).

Examples of uniquely identifiable (within a similarity transform) model parameters include, $\CovY$, $\Sigma_e$, $K$, $C_y$, and $A$. Notably, unlike the total Kalman gain $K$, its components $K_f$ and $K_v$ (equation \ref{eq:Kalman_K_def}) are \textit{not} uniquely identifiable. This has an important ramification for Kalman prediction versus Kalman filtering. While the Kalman prediction (\ref{eq:Kalman_state_prediction}) only relies on uniquely identifiable parameters (i.e., $\{A, C_y, K\}$), the update step needed for Kalman filtering (\ref{eq:Kalman_state_update}) relies on $K_f$, which is not uniquely identifiable. This means that given time series ${\y{k}}$ from a system with latent states, the optimal Kalman filter is uniquely identifiable, whereas there are infinitely many Kalman filters associated with that unique Kalman predictor that are equivalent in terms of how they describe $\y{k}$. This is also intuitively clear, because given a sample of the time series $\y{k}$, estimating that same time step given its true observed value is a trivial problem that does not require a filter: the optimal estimation of the denoised value of $\y{k}$ (i.e., $C_y \x{k}$) given $\y{k}$ (i.e., $C_y \x{k} + \vk{}$) is simply $\y{k}$ itself, which would yield the minimum possible expected error of $\vk{}$. This is because $\vk{}$ is white, and thus no amount of additional observations from other samples besides $\y{k}$ can provide any information about $\vk{}$, making it the minimum possible error. The same holds for optimal smoothing for $\y{k}$ given $\y{k}$ itself. We emphasize that this triviality of filtering/smoothing and the un-identifiability of an optimal filter/smoother is only the case for systems with \textit{latent} states, not those with measurable states. 

The above is only true when only one time series is available. In the PSID setting, where a second time series $\z{k}$ is available, the joint second order statistics of the two time series are the objective of identification, and this expanded scope disambiguates the identification problem compared with the one-signal cases. In the PSID setting, we further assume that the secondary signal is only measured during training, and only the primary signal is measured during inference. In this setting, non-trivial filtering and smoothing problems can be defined as follows: the optimal filtering is the best estimate of $\z{k}$ given all samples of $\y{k}$ up to $k$. The optimal smoothing is the best estimate of $\z{k}$ given all samples of $\y{k}$ up to $N$. In other words, in the PSID setting, the presence of a secondary time series $\z{k}$ during system identification creates a non-trivial filtering and smoothing problem for that secondary time series. As we will show here, this means that otherwise unidentifiable parameters such as $K_f$ become partially identifiable (to the extent that they are related to $\z{k}$). 

\subsection{PSID}
\label{sec:psid_main}

Preferential Subspace Identification (PSID) is a system identification method designed to model the dynamics of two time series, $\yk{} \in \mathbb{R}^{n_y}$ and $\zk{} \in \mathbb{R}^{n_z}$, the latter of which is not expected to be measured during inference. A key use-case for PSID is modeling neural-behavioral data for use in brain-machine interfaces, where the behavior signal is often a target for decoding and is not measured during inference. However, the method is general and can be applied to any pair of time series.

The key insight of PSID is to identify the dynamics of the primary signal $\yk{}$ while dissociating dynamics that are relevant to the secondary signal $\zk{}$ from those that are unrelated to $\zk{}$. PSID further prioritizes learning the dynamics that are shared between the two time series.

PSID operates in two stages. In the first stage, dynamics that are shared between the two time series are extracted via a projection of future behavior $\z{k}$ onto corresponding past neural activity $\y{k}$. In the second stage, any residual dynamics in neural activity that are not explained by the latent states extracted in the first stage are explained using additional latent states. The second stage identifies these additional states by projecting future residual activity onto past neural activity.

In the first stage, a pre-specified number of latent states, denoted by $n_1$, are extracted. In the second stage, an additional pre-specified number of latent states, denoted by $n_x - n_1$, are extracted. After all parameters are learned, the overall model takes the form of equation \ref{eq:model_psid}, which is equivalent to equation \ref{eq:stochastic_form}.

\begin{equation}
   \label{eq:model_psid}
   \left\{
   \begin{array}{cccccc}
       \begin{bmatrix} \xknextsup{1} \\ \xknextsup{2} \end{bmatrix} &=& 
       \begin{bmatrix} A_{11} & A_{12} \\ A_{21} & A_{22} \end{bmatrix} &
       \begin{bmatrix} \xksup{1} \\ \xksup{2} \end{bmatrix} &+&
       \begin{bmatrix} \wksup{1} \\ \wksup{2} \end{bmatrix}
       \\[1em]
       \yk &=& 
       \begin{bmatrix} C_y^{(1)} & C_y^{(2)} \end{bmatrix} &
       \begin{bmatrix} \xksup{1} \\ \xksup{2} \end{bmatrix} &+&
       \vk
       \\[1em]
       \zk &=& 
       \begin{bmatrix} C_z^{(1)} & C_z^{(2)} \end{bmatrix} &
       \begin{bmatrix} \xksup{1} \\ \xksup{2} \end{bmatrix} &+&
       \epsk
   \end{array}
   \right.
\end{equation}

Once the PSID model parameters are learned, at inference time, a Kalman filter per equation \ref{eq:Kalman_state_prediction} can be used to extract the latent states from the neural activity and in turn predict behavior from the latent states. The latent states can simply be multiplied by the parameter $C_z$ to obtain behavior predictions:
\begin{equation}
   \label{eq:z_prediction}
   \zh{k} = C_z \xh{k}. 
\end{equation}

\subsubsection{One-step-ahead prediction versus filtering versus smoothing in the PSID setting}

As described in section \ref{eq:system_identification}, not all model parameters are uniquely identifiable, because for some of them, there are infinitely many equivalent solutions. In the context of prediction, similar to one-signal system identification, all PSID parameters are uniquely identifiable. However, in the context of filtering in the PSID setting, there is a fundamental difference compared to the normal one-signal system identification. The difference is that, given the existence of the second time series $\z{k}$, and the fact that in the first stage of PSID we are optimizing for dynamics of $\y{k}$ that are relevant to that second time series, there is now a meaningful distinction between all the equivalent models (unlike in section \ref{sec: faurres_theorem}).

These alternative models correspond to different stochastic form models, each with their own Kalman filter and predictor. While the Kalman predictor parameter $K$ is uniquely identifiable even in the one-signal setting, the Kalman filter parameter $K_f$ is not uniquely identifiable (section \ref{eq:system_identification}). In other words, all alternative stochastic form models have the same $K$ (within a similarity transform), while they do not have the same $K_f$. Moreover, these models are not all identical in terms of behavior prediction. During training, we have access to the secondary time series $\z{k}$, and the objective of the PSID algorithm is to optimize the prediction of this secondary time series.

In the case of filtering and smoothing, the objective of the extended PSID algorithm we develop in this work is to estimate the secondary signal using the primary signal samples up to the same sample (filtering) or up to the final sample number $N$ (smoothing).

\subsubsection{PSID with filtering}

To derive PSID with optimal filtering, our key idea is to select the parameter $K_f$ among all possible solutions of system identification that yields the best filtered estimate of the secondary time series $\z{k}$.

The optimization that we want to solve is:
\begin{equation}
\label{eq:filtering_optimization}
\arg\min_{K_f} \| \zk{} - \zh{k} \|_2^2
\end{equation}
where $\zh{k} = C_z \xh{k|k}$ and $\xh{k|k}$ is computed using $K_f$. Replacing $\xh{k|k}$ from equation \ref{eq:Kalman_state_update} gives:

\begin{equation}
\label{eq:filtering_optimization_expanded}
\begin{array}{rl}
   & \arg\min_{K_f} \| \zk{} - C_z \xh{k|k} \|_2^2 \\
   = & \arg\min_{K_f} \| \zk{} - C_z (\xh{k|k-1} + K_f (\y{k} - C_y \xh{k|k-1})) \|_2^2 \\
   = & \arg\min_{K_f} \| \zk{} - \zh{k|k-1} - C_z K_f (\y{k} - \yh{k|k-1}) \|_2^2 \\
   = & \arg\min_{K_f} \| \ztilde{k|k-1} - C_z K_f \ytilde{k|k-1} \|_2^2 
\end{array}
\end{equation}

where $\ztilde{k|k-1} = \zk{} - \zh{k|k-1}$ and $\ytilde{k|k-1} = \y{k} - \yh{k|k-1}$ are residuals from  Kalman one-step-ahead prediction. The linear minimum mean squared error estimate for this optimization has a closed-form solution that gives us the optimal $C_z K_f$ as follows:
\begin{equation}
\label{eq:czkf_solution}
C_z K_f = \arg\min_{M} \| \Ztilde{} - M \Ytilde{} \|_F^2 = \Ztilde{} \Ytilde{}^T (\Ytilde{} \Ytilde{}^T)^{-1}
\end{equation}
where $\Ztilde{}$ and $\Ytilde{}$ are wide matrices, the columns of which consist of $\ztilde{k|k-1}$ and $\ytilde{k|k-1}$, respectively, for all training samples. 

In fact, obtaining $C_z K_f$ is sufficient for implementing the optimal filter for predicting $\zk{}$ from $\yk{}$, without the need to learn $K_f$ separately. To do so, we simply multiply the predicted state $\xh{k|k-1}$ by the learned $C_z K_f$, which according to the Kalman filter update equations, gives us the filtered estimation of $\zk{}$.

\begin{figure}[ht]
    \centering
    \tikzset{
        box/.style={rectangle, draw, thick, minimum width=3.2cm, minimum height=1.2cm, text centered, fill=white},
        outerbox/.style={rectangle, draw, thick, minimum width=4.2cm, minimum height=8.6cm, rounded corners, fill=green!20},
        signal/.style={text centered, text width=2.5cm},
        io/.style={text centered, text width=4.5cm},
        sum/.style={circle, draw, inner sep=0pt, minimum size=0.5cm}
    }
    \begin{subfigure}[b]{\textwidth}
        \centering
        \begin{tikzpicture}[
            node distance=2.2cm, 
            auto
        ]
         \useasboundingbox (-4, -4.5) rectangle (6.5, 2.5);
    
          \node at (-5, 2.2) {\textbf{a}};
          \node[outerbox, minimum height=7cm] (outer) at (0, -1.7) {};
          \node at (0, 1.3) {\textbf{PSID with filtering}};
    
          \node[box] (psid) at (0, 0) {\textbf{PSID}};
          \node[box] (kalman) at (0, -2) {\textbf{Kalman Predictor}};
          \node[box] (rrr) at (0, -4) {\textbf{RRR}};
   
          \node[signal] (neural) at (-4.75, 0.75) {Primary \\ signals};
          \node[signal] (behavior) at (-3.25, 0.75) {Secondary \\ signals};
    
          \node[signal] (psid_inputs) at (-4, 0) {$\y{1:N}, \z{1:N}$};
          \node[signal] (kalman_inputs) at (-4, -2) {$\y{1:N}$};
          \node[signal] (rrr_inputs) at (-4, -4) {$\ytilde{1:N}, \ztilde{1:N}$};
    
          \node[io] (psid_output) at (5, 0) {$\left\{A, C_y, C_z, K, \CovY\right\}_{forward}$};
          \node[io] (kalman_output) at (5, -2) {$\yh{1|0} \dots \yh{N|N-1}, \zh{1|0} \dots \zh{N|N-1}$ \\ (one-step-ahead estimates)};
          \node[io] (rrr_output) at (5, -4) {$\left\{C_z K_f\right\}_{forward}$ \\ and $\zh{1|1} \dots \zh{N|N}$ \\ (filtered estimates)};
    
          \draw[->, thick] (psid_inputs.east) -- (psid.west);
          \draw[->, thick] (kalman_inputs.east) -- (kalman.west);
          \draw[->, thick] (rrr_inputs.east) -- (rrr.west);
          \draw[->, thick] (psid.east) -- (psid_output.west);
          \draw[->, thick] (kalman.east) -- (kalman_output.west);
          \draw[->, thick] (rrr.east) -- (rrr_output.west);
          
          \draw[->, thick] (psid_output.south) -- (5, -1) -- (0, -1) -- (kalman.north);

        \end{tikzpicture}
        \label{fig:method_filtering_psid}
    \end{subfigure}
    
    \begin{subfigure}[b]{\textwidth}
        \centering
        \begin{tikzpicture}[node distance=2.2cm, auto]
            \useasboundingbox (-4, -1.5) rectangle (6.5, 2.5);

            \node at (-5, 1.5) {\textbf{b}};

            \node[outerbox, minimum height=1.5cm] (psid_with_filtering) at (0, 0) {\textbf{PSID with filtering}};

            \node[signal] (psid_inputs) at (-4, 0) {$\y{N:1}, \ztilde{N:1}$};

            \node[io] (psid_output) at (5, 0) {$\left\{A, C_y, C_z, K, \CovY\right\}_{backward}$ \\ \vspace{0.25em} and \vspace{0.25em} \\ $\left\{C_z K_f\right\}_{backward}$ \\ \vspace{0.5em} and \vspace{0.25em} \\ $\ztildehat{N|N} \dots \ztildehat{1|1}$};

            \draw[->, thick] (psid_inputs.east) -- (psid_with_filtering.west);
            \draw[->, thick] (psid_with_filtering.east) -- (psid_output.west);

         \end{tikzpicture}
        \label{fig:method_smoothing_psid}
    \end{subfigure}
    \caption{(\textbf{a}) Diagram of PSID with filtering. The method consists of three main steps: (1) Regular PSID learns the forward model parameters from input signals, (2) a Kalman predictor uses the learned model to make one-step-ahead predictions, and (3) Reduced Rank Regression (RRR) learns updated $C_z K_f$ parameters to produce optimal filtered estimates of the behavior signals. (\textbf{b}) Diagram of PSID with smoothing. The method first applies PSID with filtering as in (\textbf{a}), and obtains the error of the filtered estimate of the secondary signal, i.e., $\ztilde{1:N}$. Next, $\y{}$ and $\ztilde{}$ are reversed in time, i.e., $\y{N:1}$ and $\ztilde{N:1}$, and passed to PSID with filtering to learn the parameter of the backwards model.}
    \label{fig:method}
\end{figure}
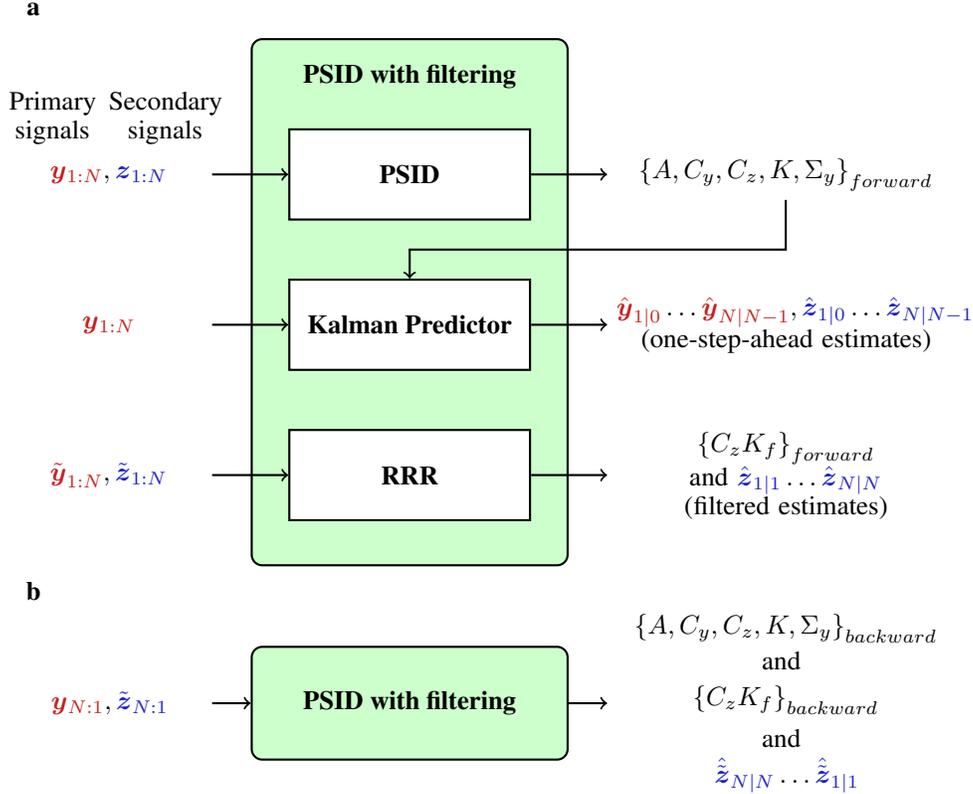

One critical point is that the linear minimum mean squared estimate noted in the equation above may not be correct if $n_y > n_x$ and $n_z > n_x$, in the sense that it may have a rank larger than $n_x$, whereas we expect the rank of $C_z K_f$ to be at most equal to $n_x$, or more precisely at most $min(n_x, n_y, n_z)$. Therefore, instead of using the linear minimum mean squared estimate, we use a reduced-rank regression (RRR) solution to enforce the rank of $C_z K_f$ to be at most $n_x$ (Figure \ref{fig:method}a).

Finally, we can use a more general version of equation \ref{eq:filtering_optimization_expanded} where we learn $\Gamma_z K_f$, where $\Gamma_z$ is the extended observability matrix for the pair $(C_z, A)$, instead of learning $C_z K_f$, as follows:

\begin{equation}
   \label{eq:filtering_optimization_expanded_observability}
   \begin{array}{rl}
      & \arg\min_{K_f} \sum_{l=0}^{i-1} \| \z{k+l} - C_z A^l \xh{k|k} \|_2^2 \\
      = & \arg\min_{K_f} \sum_{l=0}^{i-1} \| \z{k+l} - C_z (A^l\xh{k|k-1} + A^l K_f (\y{k} - C_y \xh{k|k-1})) \|_2^2 \\
      = & \arg\min_{K_f} \sum_{l=0}^{i-1} \| \z{k+l} - \zh{k+l|k-1} - C_z A^l K_f (\y{k} - \yh{k|k-1}) \|_2^2 \\
      = & \arg\min_{K_f} \sum_{l=0}^{i-1} \| \ztilde{k+l|k-1} - C_z A^l K_f \ytilde{k|k-1} \|_2^2 \\
      = & \arg\min_{K_f} \| \begin{bmatrix} \ztilde{k|k-1} \\ \ztilde{k+1|k-1} \\ \vdots \\ \ztilde{k+i-1|k-1} \end{bmatrix} - \Gamma_z K_f \ytilde{k|k-1} \|_2^2 .

   \end{array}
\end{equation}
Here, $i$ is the PSID hyperparameter called the horizon \citep{saniModelingBehaviorallyRelevant2021}, and $\Gamma_z$, i.e., the extended observability matrix for the pair $(C_z, A)$, is defined as:

\begin{equation}
\label{eq:gamma_z}
\Gamma_z = \begin{bmatrix} C_z \\ C_z A \\ \vdots \\ C_z A^{i-1} \end{bmatrix}.
\end{equation}

This more general optimization can be converted to matrix form as in equation \ref{eq:czkf_solution} by forming matrices whose columns are the terms of equation \ref{eq:filtering_optimization_expanded_observability} at different time steps. We can then solve for $\Gamma_z K_f$ using RRR as before, and take the first $n_z$ rows of $\Gamma_z K_f$ as $C_z K_f$. This more general approach has the benefit that with a large enough $i$, the rank of $\Gamma_z K_f$ is not limited by $n_z$, rather can be as large as $min(n_x, n_y)$, accommodating the full rank of $K_f$, the identification of which we will discuss in the next section.

\subsubsection{Solving for the exact solution for $K_f$}
\label{sec:exact_solution_for_kf}

Previously, we showed how $C_z K_f$ can be identified, and that solution is always available. We also explained why as far as the practical problem of predicting/filtering behavior is concerned, identifying $C_z K_f$ is sufficient and we do not need to identify $K_f$ separately. Here, we will discuss the conditions under which $K_f$ itself is also identifiable, which is not always the case. This is fundamentally because not all latent states are always relevant to behavior. More formally, the pair $(C_z, A)$ is not always observable, which means that even when we observe the secondary signal $\z{k}$, the latent states $\x{k}$ are not always fully observable. Thus, for these systems, even in the PSID setting where we observe $\z{k}$, the $K_f$ associated with certain latent states is not uniquely identifiable. For example, consider the special case of a system where $\z{k}=\y{k}$. In such a system, the PSID result would be the same as the regular subspace identification result, and the $K_f$ would thus not be uniquely identifiable.

However, in the special case where all latent states are relevant to the secondary signal (i.e., the pair $(C_z, A)$ is observable), $C_z K_f$ can be decomposed into updated $C_z$ and $K_f$ matrices. An example of this would be any time when $C_z$ has a left pseudo-inverse. In this case, multiplying the computed $C_z K_f$ by that left pseudo-inverse would give us the exact solution for $K_f$ that optimizes the filtering of the secondary signal. Similarly, in the more general formulation from the previous section, whenever $\Gamma_z$ has a left pseudo-inverse, multiplying the computed $\Gamma_z K_f$  by that left pseudo-inverse would give us the exact solution for $K_f$.

Since in general this solution is not available, in this new method, which we call \textit{PSID with filtering}, we always only learn $C_z K_f$ from the data and use that in generating our filtered estimate of the secondary signal.

\subsubsection{Smoothing with PSID}
\label{sec:smoothing_with_psid}

Inspired by the two-filter formulation for Kalman smoothers, we recognize that PSID learns the optimal filter in one direction. To apply a smoother, we further need to learn the optimal filter to predict residual data in the opposite direction. In effect, we are learning the forward and backward filters of the forward-backward smoother separately and directly from data.

More concretely, smoothing PSID proceeds as follows (Figure \ref{fig:method}b):

\begin{enumerate}
    \item Apply regular PSID with filtering, which, as explained in the previous section, consists of PSID for prediction plus reduced rank regression to learn $C_z K_f$.
    
    \item Compute the filtered estimate of the secondary time series $\zh{k|k}$ using this learned model.
    
    \item Subtract this filtered estimate from the secondary time series to find the residual secondary time series:
    \begin{equation}
        \ztilde{k|k} = \z{k} - \zh{k|k}.
        \label{eq:residual}
    \end{equation}
    
    \item Apply PSID with filtering---the exact same method of PSID plus reduced rank regression---in the opposite time direction and on the residual secondary signal $\ztilde{k|k}$ as our new secondary signal. This gives us a new model in the reverse time direction.
\end{enumerate}

The final prediction from this smoothing PSID, i.e., $\zh{k|N}$, is the sum of the predictions from the forward and backward filters:

\begin{equation}
    \zh{k|N} = \zh{k|k} + \ztildehat{k|k},
    \label{eq:smoothing_prediction}
\end{equation}

where $\zh{k|k}$ and $\ztildehat{k|k}$ are the forward and backward PSID filtered estimates for sample $k$, respectively (Figure \ref{fig:method}b). Note that this formulation resembles the forward-backward Kalman smoothing formulation in equation \ref{eq:forward_backward_smoother}, in that the final prediction is a weighted sum of the forward and backward predictions.

It should be noted that the backwards model learned in PSID smoothing is \textit{different} from the backward representation of the stochastic model (equation \ref{eq:stochastic_box}), which is explained in Appendix \ref{sec:appendix_note_backward_stochastic_model}. This is because here the backwards model is learned from \textit{residual} behavior data $\ztilde{k}$. An alternative approach that would learn the backwards stochastic model would be to simply pass the original behavior $\z{k}$ in the reverse direction to learn the backwards model using PSID. The final behavior prediction would then be the mean (instead of the sum) of the forward and backward models' behavior predictions. As we confirm in Appendix \ref{sec:appendix_note_alternative_smoothing}, this alternative approach would indeed learn the backwards stochastic form as its backwards model, but it is not as accurate in learning optimal smoothing for behavior as the method based on residual behaviors that was presented earlier.

\subsection{Evaluation metrics}
\label{sec:evaluation_metrics}

To validate the extensions of PSID developed in this work, we confirm that the learned model parameters are optimal using two types of metrics. First, we confirm that the learned model parameters match the optimal parameters that we know from ground truth simulated models. Second, we confirm that the obtained filtered and smoothed estimation of the secondary signal using the primary signal indeed reaches the optimal values that we would get from the true model that simulated the data.

Overall, we simulate 20 models with random parameters, generate random realizations from these models, and compute the above metrics across the models. To compare learned parameters with ground truth parameters for a given model, we first use the method presented in \citep{saniModelingBehaviorallyRelevant2021} to change the basis of the learned model via a similarity transform to one that is aligned with that of the true model. This does not change the learned model, but makes the learned parameters comparable to the true parameters. We then compute the Frobenius norm of the difference between the learned and true parameters, normalized by the Frobenius norm of the true parameters. We compute this metric for all main model parameters learned by the original PSID method (i.e., $A$, $C_y$, $C_z$, $K$, $\CovY$), as well as the additional $C_z K_f$ parameter learned in this work for PSID with filtering.

To compare the performance for the estimation of the secondary signal using the primary signal (i.e., decoding), we use a test set separate from the data used for learning the model parameters. In this test set, we find the estimated values for the secondary signal (using prediction, filtering, or smoothing) both via the learned model parameters as well as the true model parameters. We then compute the coefficient of determination (R2) between the predicted and true time series of the secondary signal in each case.

\section{Results}

\subsection{Validation of PSID with filtering}

As noted in section \ref{sec:evaluation_metrics}, we simulated 20 random models and for each model we performed PSID with filtering to learn an initial PSID model plus a reduced rank regression solution that gives us $C_z K_f$. 

\begin{figure}[ht]
    \centering
    \begin{tikzpicture}
        \node[anchor=south west,inner sep=0] (image) at (0,0) {\includegraphics[width=0.45\linewidth, trim={2.35cm 1cm 7cm 1.39cm},clip]{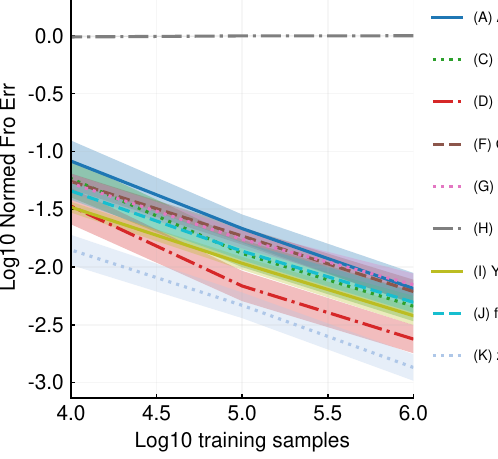}};
        \node[fill=white, text=black, anchor=south, yshift=-1mm, font=\footnotesize] at (image.south) {$Log_{10}$ of training samples};
        \node[fill=white, text=black, anchor=center, rotate=90, xshift=3mm, font=\footnotesize] at (image.west) {$Log_{10}$ Identification Error};
        \node[fill=white, anchor=west, font=\footnotesize] at (5.9, 5.55) {$A_{}$};
        \node[fill=white, anchor=west, font=\footnotesize] at (5.9, 5.019) {$C_y$};
        \node[fill=white, anchor=west, font=\footnotesize] at (5.9, 4.488) {$C_z$};
        \node[fill=white, anchor=west, font=\footnotesize] at (5.9, 3.956) {$\CovXsY$};
        \node[fill=white, anchor=west, font=\footnotesize] at (5.9, 3.425) {$K$};
        \node[fill=white, anchor=west, font=\footnotesize] at (5.9, 2.894) {$K_f$};
        \node[fill=white, anchor=west, font=\footnotesize] at (5.9, 2.363) {$\CovY$};
        \node[fill=white, anchor=west, font=\footnotesize] at (5.9, 1.831) {$C_z K_f$};
        \node[fill=white, anchor=west, font=\footnotesize] at (5.9, 1.3) {Eigs of $A$};
    \end{tikzpicture}
    \caption{The learned parameters, including the $C_z K_f$ learned for filtering, converge to the ground truth values with increasing training samples. The error for each parameter is computed as the Frobenius norm of the difference between the learned and true value, normalized by the Frobenius norm of the true value of that parameter matrix. Solid lines show the mean error across the 20 simulated models, and shaded areas show the standard error of the mean (s.e.m.). For all identifiable parameters, the mean error converges to below 1\% with 1 million training samples.}
    \label{fig:czkf_convergence}
\end{figure}
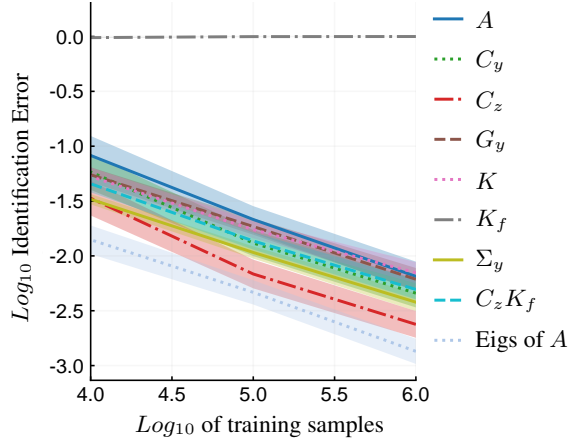

For all parameters of PSID with filtering, including the $C_z K_f$ parameter, as the number of training samples increases, the error converges to smaller and smaller values (Figure \ref{fig:czkf_convergence}). Specifically, in this simulation, with a million training samples, the average normalized error for all identifiable parameters converges to below 1\%. 

$K_f$ is an example of a non-identifiable parameter, for which as expected the error does not converge to zero (Figure \ref{fig:czkf_convergence}). For the random models in this simulation, state and observation dimensions were chosen randomly, so for many systems $n_z$ < $n_x$ and the pair $(C_z, A)$ was not observable, which means that $K_f$ was not uniquely identifiable (see section \ref{sec:exact_solution_for_kf}). Note that even though $K_f$ is an internal characteristic and not in general learnable (section \ref{sec:exact_solution_for_kf}), $C_z K_f$ which is relevant for filtering is accurately learned (Figure \ref{fig:czkf_convergence}). 

\subsection{Validation of filtering and smoothing PSID in terms of estimating behavior}

Next, we used the learned models to get filtered estimates of the secondary signal $\z{k}$. We compared these filtered estimates with the true secondary signal in the test set and computed the coefficient of determination (R2) between the two.

\begin{figure}
    \centering
    \begin{subfigure}[b]{0.3\textwidth}
        \centering
        \begin{tikzpicture}
            \node[anchor=south west,inner sep=0] (image) at (0,0) {\includegraphics[width=\linewidth, trim={2.35cm 1.5cm 8.5cm 1cm},clip]{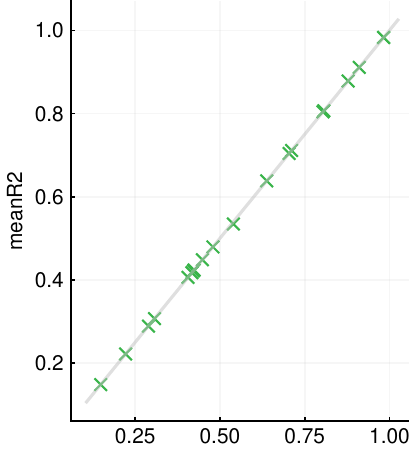}};
            \node[fill=white, text=black, anchor=south, xshift=+5mm, yshift=-4.5mm, font=\footnotesize] at (image.south) {Ideal (R2)};
            \node[fill=white, text=black, anchor=center, rotate=90, xshift=2mm, yshift=-1mm, font=\footnotesize] at (image.west) {Learned PSID model (R2)};
        \end{tikzpicture}
        \caption{Prediction}
        \label{fig:decoding_performance_pred}
    \end{subfigure}
    \hfill
    \begin{subfigure}[b]{0.3\textwidth}
        \centering
        \begin{tikzpicture}
            \node[anchor=south west,inner sep=0] (image) at (0,0) {\includegraphics[width=\linewidth, trim={2.35cm 1.5cm 8.5cm 1cm},clip]{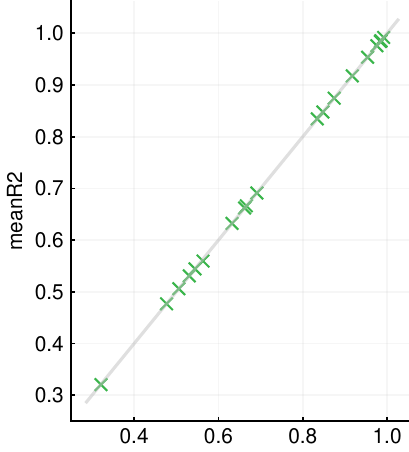}};
            \node[fill=white, text=black, anchor=south, xshift=+5mm, yshift=-4.5mm, font=\footnotesize] at (image.south) {Ideal (R2)};
            \node[fill=white, text=black, anchor=center, rotate=90, xshift=2mm, yshift=-1mm, font=\footnotesize] at (image.west) {Learned PSID model (R2)};
        \end{tikzpicture}
        \caption{Filtering}
        \label{fig:decoding_performance_filt}
    \end{subfigure}
    \hfill
    \begin{subfigure}[b]{0.3\textwidth}
        \centering
        \begin{tikzpicture}
            \node[anchor=south west,inner sep=0] (image) at (0,0) {\includegraphics[width=\linewidth, trim={2.35cm 1.5cm 8.5cm 1cm},clip]{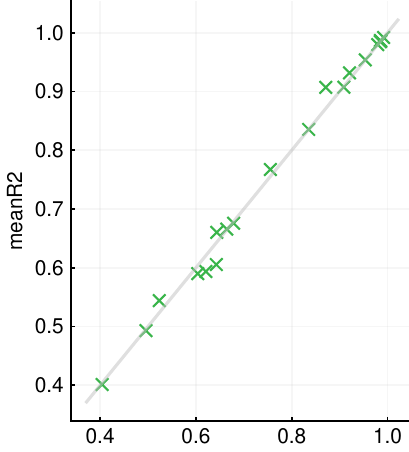}};
            \node[fill=white, text=black, anchor=south, xshift=+5mm, yshift=-4.5mm, font=\footnotesize] at (image.south) {Ideal (R2)};
            \node[fill=white, text=black, anchor=center, rotate=90, xshift=2mm, yshift=-1mm, font=\footnotesize] at (image.west) {Learned PSID model (R2)};
        \end{tikzpicture}
        \caption{Smoothing}
        \label{fig:decoding_performance_smth}
    \end{subfigure}
    \caption{Estimation performance of the secondary signal for (\textbf{a}) One-step-ahead prediction ($k|k-1$). (\textbf{b}) Filtering ($k|k$). (\textbf{c}) Smoothing ($k|N$), for true models versus the models learned using our extended PSID method. Each point represents one simulated model with random parameters. The horizontal axis shows the performance of the true model and the vertical axis shows the performance of the learned model.}
    \label{fig:decoding_performance}
\end{figure}
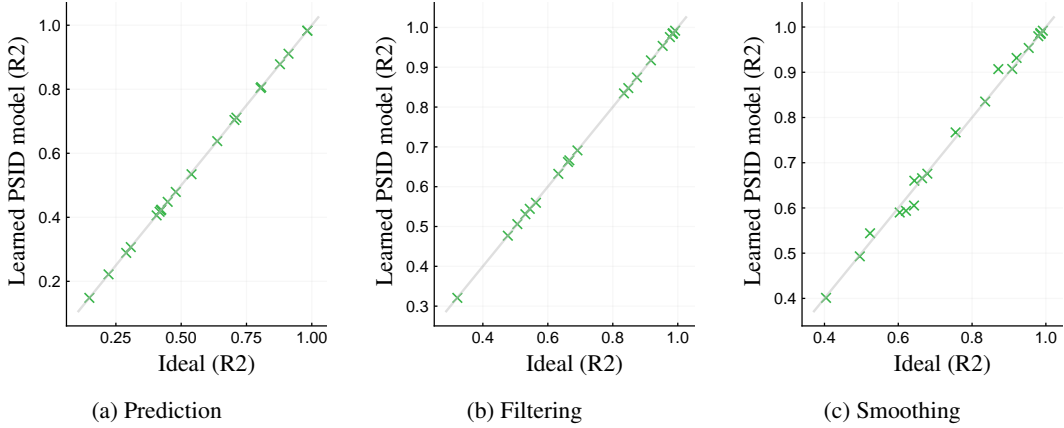

We also did the same for the true models. That is, we used the true model to perform filtering to get filtered estimates of the secondary signal in the test set from the primary signal. As we see in Figure \ref{fig:decoding_performance}b, the filtered estimate of the secondary signal is similar to the performance of the true models. The results are similar to those obtained for the 1-step ahead predictions obtained from the original PSID (figure \ref{fig:decoding_performance}a).

Similarly, we used our learned models to perform smoothing to find the smoothed estimate of the secondary signal from the complete time samples of the primary signal. We also did the same using the true models and then computed the R2 between the smoothed estimate of the secondary signal in each case with the true secondary signal in the test set. As we see in Figure \ref{fig:decoding_performance}c, the smoothed estimate of the secondary signal is similar to the performance of the true models, confirming that the learned models also achieve optimal smoothing of the secondary signal.

\section{Discussion}

Here, we developed extensions of PSID that enable the optimal filtering or smoothing of a secondary signal using a primary time series. We show with connections to fundamental system identification concepts that having a secondary signal creates a profound difference in terms of which internal model parameters are uniquely identifiable and which internal parameters are not.

To recap, in the single signal system identification setup with latent states, there is no fundamental difference between different stochastic form models because they are all equivalent. The concepts of filtering and smoothing are not interesting, because the optimal version of them is simply predicting the observed signal at that sample as itself. However, in the PSID setup, this fundamentally changes. Now, the secondary signal is our metric for determining which of the equivalent stochastic form models is better. The stochastic form model that yields the best filtered or smoothed estimation of the secondary signal is optimal. So of all equivalent stochastic form solutions, only one of them would apply here.

However, despite this additional visibility into the internals of the system that is afforded to us by having a secondary time series, we cannot learn all internal parameters uniquely. This is because not all of the internal parameters affect the secondary signal. In cases where they do, we explained how the exact $K_f$ can be identified. Identifying the associated $Q$, $R$, $S$ noise statistics for the stochastic form that give a particular $K_f$ is an interesting follow-up problem that we did not tackle here. 

Regarding PSID with filtering, one natural question is whether one could have achieved the same optimal filtering accuracy by simply shifting the secondary signal one sample forward in time during training and then applying the original PSID algorithm. In Appendix~\ref{sec:appendix_note_shifting} we show that while this "shifted PSID" baseline indeed improves performance over ideal one-step-ahead prediction, it still falls short of optimal filtering. We also explain theoretically why correlations between the state and observation noises make this "shifted PSID" approach suboptimal. In contrast, as we confirm in simulations, the PSID with filtering method presented here reaches optimal filtering regardless of noise correlations (Figure \ref{fig:decoding_performance_filt}).

The theoretical results in this work are validated through various numerical simulations, which demonstrate the optimality of the proposed PSID with filtering/smoothing methods for decoding the secondary signal from the primary signal. While this paper focuses on the methodological development, the primary and secondary signals can be any two time series, and thus the method here can be used in developing various applications and solving different problems such as those in neuroscience. For example, our concurrent work in \citet{jhaPrioritizedLearningCrosspopulation2025} formulates the problem of identifying cross-regional neural dynamics as a prioritized learning problem, thus enabling the utilization of the methods here for solving that problem.

Finally, the similar concept to what was used here---learning a forward pass model and learning a backward pass model on the residual---is also applicable in more general non-linear decoding settings, such as those addressed by DPAD \citep{saniDissociativePrioritizedModeling2024}.

\begin{ack}
We sincerely thank Trisha Jha for giving feedback on drafts of this manuscript.

This work was supported, in part, by the following organizations and grants: the Office of Naval Research (ONR) Young Investigator Program under contract N00014-19-1-2128, National Institutes of Health (NIH) Director's New Innovator Award DP2-MH126378, NIH R01MH123770, NIH BRAIN Initiative R61MH135407 and the Army Research Office (ARO) under contract W911NF-16-1-0368 as part of the collaboration between the US DOD, the UK MOD and the UK Engineering and Physical Research Council (EPSRC) under the Multidisciplinary University Research Initiative (MURI).

\end{ack}
    
{
\small
\bibliographystyle{plainnat}
\bibliography{references}
}


\appendix

\section{Appendix}

\subsection{Backward stochastic model}
\label{sec:appendix_note_backward_stochastic_model}

Equivalent to the stochastic form representation of a model (section \ref{sec:stochastic_vs_predictor}), one can also describe the same second order statistics of the observations in terms of a backward stochastic model where the direction of time is reversed \citep{vanoverscheeSubspaceIdentificationLinear1996}. Specifically, this formulation, which is repeated below, is referred to as the backward stochastic model:

\begin{tcolorbox}[title=The backward stochastic form, halign title=center]
    \centering
    \begin{subequations}
    \label{eq:backward_stochastic_box}
    \begin{align}
    \xb{k-1} &= A^T \xb{k} + \wbk{} \\
    \y{k} &= \CovXsY^T \xb{k} + \vbk{} \\
    \mathbb{E}\left(\begin{bmatrix} \wb{p} \\ \vb{p} \end{bmatrix} \begin{bmatrix} \wb{q} \\ \vb{q} \end{bmatrix}^T\right) &= \begin{pmatrix} Q^b & S^b \\ (S^b)^T & R^b \end{pmatrix} \delta_{pq}
    \end{align}
    \end{subequations}
    \begin{subequations}
    \label{eq:stochastic_box_covariances}
    \begin{align}
    \mathbb{E}[\xb{k} (\xb{k})^T] &\triangleq (\CovXs)^{-1} = A^T (\CovXs)^{-1} A + Q^{b}, \\
    \mathbb{E}[\y{k} \y{k}^T] &\triangleq \CovY = \CovXsY^T (\CovXs)^{-1} \CovXsY + R^{b}, \\
    \mathbb{E}[\xb{k-1} \y{k}^T] &\triangleq (C_y)^T = A^T (\CovXs)^{-1} \CovXsY + S^{b}.
    \end{align}
    \end{subequations}
\end{tcolorbox}
    
Here, the latent state of the backwards model $\xb{k}$ is related to that of the forward model, i.e., $\xs{k}$, by the following relationship \citep{vanoverscheeSubspaceIdentificationLinear1996}:
\begin{equation}
    \label{eq:backward_state_relationship}
    \xb{k} \triangleq \CovXs^{-1} \xs{k}
\end{equation}

Additional derivations for the other relations between the backward and forward stochastic models are provided in \citet{vanoverscheeSubspaceIdentificationLinear1996}. What we need to add here is the readout equation for the secondary signal $\z{k}$ in the backward model. Before we derive this --- similar to how the primary readout is derived in \citet{vanoverscheeSubspaceIdentificationLinear1996} --- we will recall the readout equation for the secondary signal $\z{k}$ (equation \ref{eq:stochastic_form}) in the forward stochastic model:
\begin{equation}
    \label{eq:secondary_readout}
    \zk{} = C_z \xs{k} + \epsk{} 
    .
\end{equation}
 
We will also need to compute the cross-covariance between the secondary signal $\z{k}$ and the forward latent state $\xs{k+1}$, denoted by $\CovXsZ$, as follows:
\begin{subequations}
    \begin{align}
        \label{eq:secondary_cross_covariance}
        \CovXsZ \triangleq \mathbb{E}[\xs{k+1} \z{k}^T] &= \mathbb{E}[(A \xs{k} + \wk{}) (C_z \xs{k} + \epsk{})^T] \\
        &= A \CovXs C_z^T + S_{xz}
    \end{align}
\end{subequations}
 
where $S_{xz} \triangleq \mathbb{E}[\wk{}\veps{}_k^T]$. Finally, we denote the minimum variance estimate of one random variable given the other as $\Pi(.|.)$. 

We can then derive the readout equation for the secondary signal $\z{k}$ in the backward model as follows:
\begin{subequations}
\begin{align}
    \z{k} &= \Pi(\z{k} \mid \xs{k+1}) + (\z{k} - \Pi(\z{k} \mid \xs{k+1})) \\
          &= \mathbb{E}[\z{k} (\xs{k+1})^T](\mathbb{E}[\xs{k+1}(\xs{k+1})^T])^{-1}\xs{k+1} + (\z{k} - \Pi(\z{k} \mid \xs{k+1})) \\
          &= \mathbb{E}[(C_z \xs{k} + \epsk{})((\xs{k})^T A^T + \wk{}^T)] \, \CovXs^{-1} \xs{k+1} + (\z{k} - \Pi(\z{k} \mid \xs{k+1})) \\
          &= (C_z \CovXs A^T + S_{xz}^T) \, \CovXs^{-1} \, \xs{k+1} + (\z{k} - \Pi(\z{k} \mid \xs{k+1})) \\
          &= \CovXsZ^T \xb{k} + \epsbk{}
    \label{eq:backward_output_derivation}
\end{align}
\end{subequations}
where $\epsbk{} \triangleq \z{k} - \Pi(\z{k} \mid \xs{k+1})$. 

For simplicity in presenting metrics for the learning of the backward model parameters in Appendix \ref{sec:appendix_note_alternative_smoothing}, we will denote each parameter of the backward stochastic model as the same symbol as in the forward stochastic model, but in curly braces with a \textit{bw} subscript:

\begin{subequations}
    \begin{alignat}{3}
        \label{eq:backward_parameters}
        \{A\}_{bw} &\triangleq A^T, \qquad & 
        \{C_y\}_{bw} &\triangleq \CovXsY^T, \qquad & 
        \{C_z\}_{bw} &\triangleq \CovXsZ^T, \\
        \{\CovXsY\}_{bw} &\triangleq C_y^T, \qquad & 
        \{\CovY\}_{bw} &\triangleq \CovY, \qquad & 
        &
    \end{alignat}
\end{subequations}

and the Kalman gain parameters are computed per equation \ref{eq:Kalman_gain_defs}, but based on the above backwards parameters.

\subsection{Alternative backward model for PSID smoothing}
\label{sec:appendix_note_alternative_smoothing}

As noted in the main text, the backward model learned for smoothing in our proposed method is different from the backward representation of the underlying stochastic model as formulated in Figure 3.5 of \citet{vanoverscheeSubspaceIdentificationLinear1996} and Appendix \ref{sec:appendix_note_backward_stochastic_model}. 
We empirically demonstrate this distinction here. The primary difference lies in the data used to train the backward model. In our proposed PSID smoothing approach, the backward model is learned using the \textit{residual} of the secondary signal $\ztilde{k}$, i.e., the portion of the secondary signal not explained by the forward PSID model. An alternative approach would be to train the backward model on the time-reversed secondary signal $\z{k}$ itself (see section \ref{sec:smoothing_with_psid}). As we validate here, this alternative procedure indeed learns the backward stochastic model from \citet{vanoverscheeSubspaceIdentificationLinear1996} (see Appendix \ref{sec:appendix_note_backward_stochastic_model}).

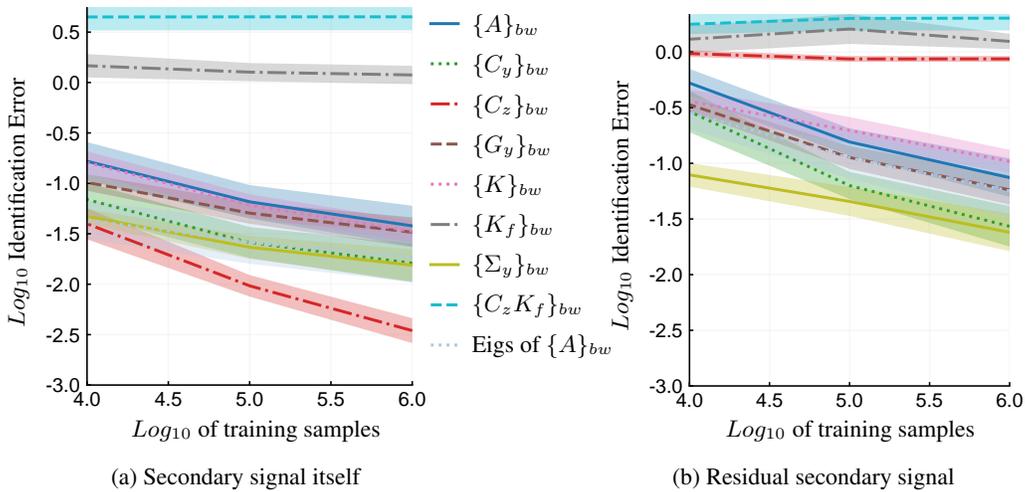
\begin{figure}[ht]
    \centering
    \begin{subfigure}[b]{0.45\textwidth}
        \centering
        \begin{tikzpicture}
            \node[anchor=south west,inner sep=0] (image) at (0,0) {\includegraphics[width=1\linewidth, trim={2.35cm 1cm 7cm 1.39cm},clip]{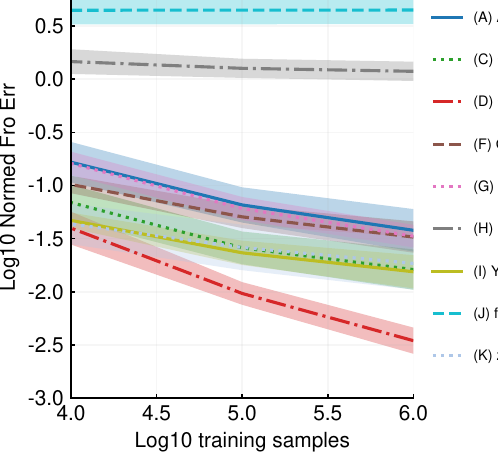}};
            \node[fill=white, text=black, anchor=south, yshift=-1mm, font=\footnotesize] at (image.south) {$Log_{10}$ of training samples};
            \node[fill=white, text=black, anchor=center, rotate=90, xshift=3mm, font=\footnotesize] at (image.west) {$Log_{10}$ Identification Error};
            \node[fill=white, anchor=west, font=\footnotesize] at (5.9, 5.55) {$\{A\}_{bw}$};
            \node[fill=white, anchor=west, font=\footnotesize] at (5.9, 5.019) {$\{C_y\}_{bw}$};
            \node[fill=white, anchor=west, font=\footnotesize] at (5.9, 4.488) {$\{C_z\}_{bw}$};
            \node[fill=white, anchor=west, font=\footnotesize] at (5.9, 3.956) {$\{\CovXsY\}_{bw}$};
            \node[fill=white, anchor=west, font=\footnotesize] at (5.9, 3.425) {$\{K\}_{bw}$};
            \node[fill=white, anchor=west, font=\footnotesize] at (5.9, 2.894) {$\{K_f\}_{bw}$};
            \node[fill=white, anchor=west, font=\footnotesize] at (5.9, 2.363) {$\{\CovY\}_{bw}$};
            \node[fill=white, anchor=west, font=\footnotesize] at (5.9, 1.831) {$\{C_z K_f\}_{bw}$};
            \node[fill=white, anchor=west, font=\footnotesize] at (5.9, 1.3) {Eigs of $\{A\}_{bw}$};
            \end{tikzpicture}
        \caption{Secondary signal itself}
        \label{fig:supp_secondary_signal}
    \end{subfigure}
    \hfill
    \begin{subfigure}[b]{0.45\textwidth}
        \centering
        \begin{tikzpicture}
            \node[anchor=south west,inner sep=0] (image) at (0,0) {\includegraphics[width=0.85\linewidth, trim={2.35cm 1cm 8.15cm 1.39cm},clip]{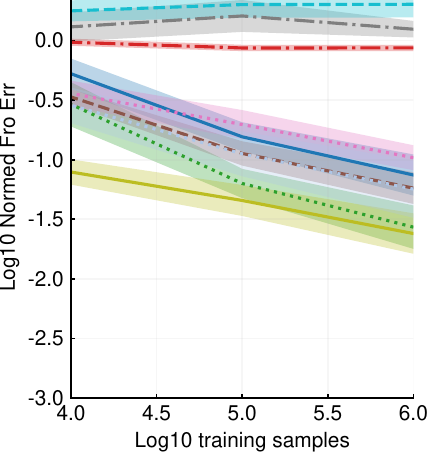}};
            \node[fill=white, text=black, anchor=south, xshift=3.5mm, yshift=-1mm, font=\footnotesize] at (image.south) {$Log_{10}$ of training samples};
            \node[fill=white, text=black, anchor=center, rotate=90, xshift=3mm, font=\footnotesize] at (image.west) {$Log_{10}$ Identification Error};
        \end{tikzpicture}
        
        \caption{Residual secondary signal}
        \label{fig:supp_residual_signal}
    \end{subfigure}
    \caption{The relation between the learned backward PSID model and the backward stochastic form of the state space model. The normalized difference between the learned parameters of the backwards model and the backward stochastic form when (\textbf{a}) the secondary signal itself is used to learn the backward model, or (\textbf{b}) the residual secondary signal is used to learn the backward model. The latter case as expected is further away from the backward stochastic form.}
    \label{fig:supplementary_smoothing_backward_methods}
\end{figure}

Figure~\ref{fig:supplementary_smoothing_backward_methods} presents an empirical comparison of these two alternative backward passes. We simulated data from models with random parameters, and compared the learned parameters for the backward PSID model with the parameters of the backward stochastic form representation of the true model. Figure~\ref{fig:supp_secondary_signal} shows the difference between the parameters learned with the alternative method (using secondary signal itself in the backward pass) with the parameters of the backward stochastic form. The identified parameters indeed converge increasingly closer to the parameters of the backward stochastic form. In contrast, as shown in Figure~\ref{fig:supp_residual_signal}, our proposed method, which uses the residuals of the forward filter, learns a different backward model (as expected) that is further away from the backward stochastic form. This model is tailored to explaining the errors of the forward pass, leading to superior (as high as ideal) smoothing performance as shown in the main text (Figure \ref{fig:decoding_performance_smth}).

\subsection{PSID cannot be extended to filtering by just shifting the training behavior data}
\label{sec:appendix_note_shifting}

Ultimately, PSID is optimizing the one-step-ahead prediction of the secondary signal using the primary signal. One might ask: if we shift the behavior signal one step forward in time during training, wouldn't that simply result in optimal filtering of the secondary signal? This is an interesting idea, and indeed it does improve the filtering performance of the secondary signal using PSID. However, as we show in this section, the optimal filter in the general case where state and observation noises are correlated (that is, $S \neq 0$ in equation~\ref{eq:def_qrs}) is not a simple shifted predictor and requires a two-step filtering and update procedure during filtering.

\begin{figure}[h]
    \centering
    \begin{tikzpicture}
        \node[anchor=south west,inner sep=0] (image) at (0,0) {\includegraphics[width=0.4\textwidth, trim={1cm 6.1cm 10cm 1cm}, clip]{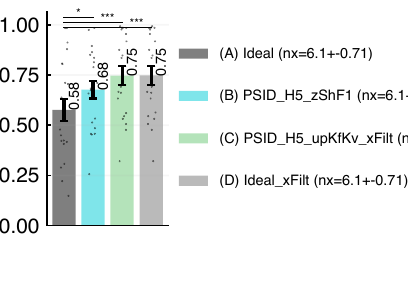}};
        \node[fill=white, text=black, anchor=south, xshift=-1.3cm, yshift=-1mm, font=\footnotesize] at (image.south) {(A) (B) (C) (D)};
        \node[fill=white, text=black, anchor=center, rotate=90, xshift=3mm, yshift=0.25cm, font=\footnotesize] at (image.west) {Estimation R2};
        \node[fill=white, anchor=west, font=\footnotesize] at (3, 3.05) {(A) Ideal prediction};
        \node[fill=white, anchor=west, font=\footnotesize] at (3, 2.519) {(B) Prediction with shifted data};
        \node[fill=white, anchor=west, font=\footnotesize] at (3, 1.988) {(C) PSID with filtering};
        \node[fill=white, anchor=west, font=\footnotesize] at (3, 1.456) {(D) Ideal filtering\hspace{0.5cm} };
    \end{tikzpicture}
    \caption{Comparison of the performance of the shifted PSID and PSID with filtering. While PSID with shifted data outperforms even an ideal (ground truth) 1-step ahead prediction, it does not reach ideal filtering accuracy, whereas the new PSID with filtering method reaches ideal filtering accuracy.}
    \label{fig:supp_shift}
\end{figure}
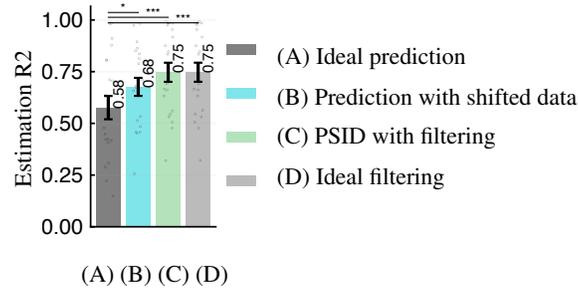

We also empirically compare the decoding performance (R2) for the shifted PSID as a baseline and show that while this baseline does improve the estimation accuracy over ideal one-step-ahead prediction, it does not reach the filtering performance of the true models, whereas PSID with filtering does achieve optimal performance (Figure~\ref{fig:supp_shift}).

We can see why this shifted-data PSID approach cannot reach optimal filtering by inspecting the Kalman filter equations (section~\ref{sec:kalman_filter}). The original PSID method identifies the parameters of the predictor form of a state-space model, including the predictor gain $K$ (equation~\ref{eq:Kalman_K_def}). This is sufficient for one-step-ahead prediction (equation~\ref{eq:Kalman_state_prediction}). However, optimal filtering requires the update step in equation~\ref{eq:Kalman_state_update}, which uses the filter gain $K_f$. As shown in equation~\ref{eq:Kalman_K_def}, the total gain is $K = A K_f + K_v$. When $S \neq 0$, $K_v$ (equation~\ref{eq:Kalman_Kv_def}) is non-zero, and thus $K_f$ cannot be uniquely determined from the predictor parameters $A$ and $K$. Since the standard PSID procedure does not identify $K_f$, it cannot produce an optimal filtered estimate in the general case.


\end{document}